# Regularized Deep Networks in Intelligent Transportation Systems: A Taxonomy and a Case Study

**Mohammad Mahdi Bejani[1], Mehdi Ghatee[2]**

1-Department of Computer Science, Amirkabir University of Technology, Iran, Tehran
2- Department of Computer Science, Amirkabir University of Technology, Iran, Tehran

**Abstract**

Intelligent Transportation Systems (ITS) are much correlated with data science mechanisms. Among the different correlation branches, this paper focuses on the neural network learning models. Some of the considered models are shallow and they get some user-defined features and learn the relationship, while deep models extract the necessary features before learning by themselves. Both of these paradigms are utilized in the recent intelligent transportation systems (ITS) to support decision-making by the aid of different operations such as frequent patterns mining, regression, clustering, and classification. When these learners cannot generalize the results and just memorize the training samples, they fail to support the necessities. In these cases, the testing error is bigger than the training error. This phenomenon is addressed as overfitting in the literature. Because, this issue decreases the reliability of learning systems, in ITS applications, we cannot use such over-fitted machine learning models for different tasks such as traffic prediction, the signal controlling, safety applications, emergency responses, mode detection, driving evaluation, etc. Besides, deep learning models use a great number of hyper-parameters, the overfitting in deep models is more attention. To solve this problem, the regularized learning models can be followed. The aim of this paper is to review the approaches presented to regularize the overfitting in different categories of ITS studies. Then, we give a case study on driving safety that uses a regularized version of the convolutional neural network (CNN).

---

[1] Ph.d. Candidate of Computer Science, mbejani@aut.ac.ir
[2] Associate Professor of Computer Science, ghatee@aut.ac.ir

## 1-Overfitting Paradigms

Choosing a model of a machine learning suitable to a dataset is a challenging problem. A simple model usually cannot solve the problem that is referred as **underfitting** [1] and a complex model memorizes the training data and cannot generalize the results for new data that is addressed as **overfitting** [1]. In both conditions, since the model cannot recognize different unseen input data, the learning process fails. The simplest way to solve the under-fitting problem is extending a more complex model with more hyper-parameters or non-linearity. But, the overfitting cannot be solved simply. The overfitting problem can be detected in the following cases:
- Great number of model parameters [2,3],
- Existing noise in training dataset [4],
- Lack of samples in training data-set (under-sampled training data)[5,6],
- Biased training samples or disproportionate training data sampling,
- Terminating the learning algorithm rapidly without convergence or dropping in a local minimum [7].

For solving the overfitting problem, many schemes were proposed to prevent from memorizing the training data. The deep networks have dramatically performance on different datasets because these models have many parameters. In addition, this property causes a disadvantage, overfitting. Then, a main problem of the deep model is overfitting. In the following, we review some of the popular schemes for controlling the overfitting in deep model named regularization schemes. We categorize the regularization schemes to two groups, error function and model based regularization.

### 1-1 Error Function based Regularization

This method has been proposed by Tikhonov and Arsenin [8]. The goal of learning is that predicts the unknown samples correctly. For this purpose, we train our model on a samples of training dataset and decrease the error of model on this data-set named empirical error that is shown in equation (1).

$$\min_{f} E\left(f; \{x_i, y_i\}_{i=1}^{D}\right), \tag{1}$$

where, $f$ is real value function that map input space to output space, $\{x_i, y_i\}_{i=1}^{D}$ is the training data-set with length $D$. The decreasing the empirical error may not cause to decrease the original error of the model. The complexity of model is one the causes. Then, we add a term to empirical error to control the complexity of the model as following.

$$min_f \; E^*(f; \{x_i, y_i\}_{i=1}^{D}) = E(f; \{x_i, y_i\}_{i=1}^{D}) + \lambda \, R(f), \qquad (2)$$

where $R(f)$ is function so that if $f$ is close to linear function the value of $R(f)$ closes to 0 and if $f$ is complex (non-linear) function, the value of $R(f)$ is great. This function is shown in equation (3). When $m = 1$, the summation of the values of learning function id considered. When $m = 2$, $R(f)$ shows the summation of changes of the learner function.

The coefficient of $\lambda$ has regularization effect, which determines the importance of the model complexity. This coefficient is gained by try and error procedure. By solving the problem (2), in optimal condition, we find the simplest function, which minimizes the empirical error. This function does not memorize the training dataset then the overfitting does not occur.

$$R(f) = \int_X ||\frac{\partial^m f}{\partial^m x}|| dx \qquad (3)$$

## 2-1 Model based Regularization

These types of schemes implicitly affect the models and the train dataset. As following, we describe some the important schemes for model based regularization.

### 2-1-1 Dropout

The dropout [7] is a most popular scheme, which controls the contribution of neurons in training process. That means, in each epoch of training process by a Bernoulli probability, $p$, the weights of neurons are trained or not. In other approach, the noise impose to inputs. The constant noise may make the model over-fitted because in training process the model learned this noise as a principal part of input data. Even thought, by imposing the random noise to input data in each iteration of learning process, we do not allow the model that learns the constant noise of the input data. There are many other schemes such as [9, 10, 11, 12, 13], which add random noise to input for regularization.

**2-1-2 Augmentation**

As we mentioned, the size of dataset is important criteria, which causes overfitting. The augmentation of the dataset is a suitable way to increase size of dataset samples. As some instances, in [14, 15, 16, 17], the experimental results of affine transition as augmentation scheme were shown. Also, in [18], the effect of adding noise to input of models as augmentation was shown. In Table

Another way to increase the number of dataset is generative models. Generative Adversarial Network [19] is a type of deep network which its' purpose is to generate new examples. In this type of models, two learning models are considered.
- This model is a map from a noise variable $p_z(z)$ to data space which is shown by $G(z, \theta_g)$ with parameter $\theta_g$.
- This model which is represented by $D(x, \theta_d)$, calculate the probability that $x$ came from the training data rather than $p_z$.

For training these models, the $\theta_g$ and $\theta_d$ should be determined so that $D(x)$ can classify $x$. For training a GAN model, the following problem should be solved:

$$min_G max_D V(D, G) = \mathbb{E}_{x \sim p_{Data}(x)} [log\, D(x)] + \mathbb{E}_{z \sim p_z(z)}[log\, (1 - D(G(z)))] \qquad (4)$$

**2-Overfitting in ITS studies**

In this section, we discuss overfitting in the data science field of ITS, especially in supervise deep models. Also, we bring two examples of the application of convolution neural network (CNN) in ITS and compare with shallow models. A suitable survey is provided by Zhang and et al. [20] for data science applications in ITS.

Having amazing performance of deep models in the computer vision and signal processing [21] causes to use these models in the different fields of engineering. When the features of a dataset could not be extracted manually, we should allow the learning model to discover the input space for extracting necessary features.

This happens when the input space of the dataset is very complex such as sound classification datasets.

When the environment of a problem is very complex, we cannot extract all of the important features. In addition, if some of these features are eliminated, there is no guarantee that the model works currently in real environments. Convolution Neural Network (CNN) is a kind of deep networks, which can extract these features automatically. The CNN by using shared weight layer extracts feature through the learning process. In [22], Szegedy et al. proposed the architecture of a network namely GoogleNet by using the power of convolution layers. Many different types of convolution in GoogleNet, cause to extract the different aspects of raw data which are hard to extract them manually. There are many other networks were proposed such as [23, 24] that by using the suitable connection between layers extract some important features. Therefore, CNNs as feature extractor and classifier is a state-of-the-art model. But, there are some problems to train a CNN.

- CNNs have many parameters, which causes to very complex loss function. Then, we need some special optimization algorithm to reduce the level of the loss function. There were proposed some algorithms, which are reviewed in [25].
- The overfitting is one of the biggest problems in machine learning. There are many causes that overfitting occurs such as noisy dataset, insufficient records in dataset, unbalance dataset, or complex model. The datasets in ITS are noisy, insufficient records, or unbalance. For example, in [26], the authors indicate that for training a multilayer perceptron model, we need to use a large amount of training data. Nevertheless, the cost of collecting data is high, then we use another approach to decrease the overfitting level namely regularization.

The other papers in ITS, which faced with overfitting phenomenon, are shown in Table 1.

*Table 1: Overfitting Challenge in ITS paper*

| Ref. | Application | Mentioned Regularization Scheme | Reported Regularization Experiments |
|------|-------------|--------------------------------|-------------------------------------|
| [28] | Using some different shallow models such as K-NN, SVM, DT, Bag, and RF for classify the transportation modes to car, bicycle, bus, walking, and running. The data-set is collecting from accelerometer, gyroscope, and rotation sensors of smartphones | The regularization method is used on DT named Cost Complexity Pruning or Weakest Link Pruning. In addition, by changing the hyper parameters of the used SVM model the authors deal with overfitting. | Yes |
| [29] | Using different shallow models for forecasting traffic flow. Loop detector sensors collect the dataset. | For regularization in neural network, this paper suggests Early Stopping and Tikhonov methods. | Yes |
| [30] | Proposed a system for forecasting traffic condition based on expectation-maximization (EM) algorithm. The input data is the GPS data of different cars in the network. | By considering different size of the input | No |
| [31] | The traffic flow data is converted to image and by using a CNN, the traffic flow is predicted. The traffic image is $N \times Q$ matrix, which $N$, $Q$ are time-segment and road segment, respectively. Each element, $i^{th}$ row and $j^{th}$ column, of the matrix shows the average speed in $i^{th}$-time and $j^{th}$-road. | The early stopping scheme is used | No |

*Table 2: Continue of Table 1*

| Ref. | Application | Mentioned Regularization Scheme | Reported Regularization Experiments |
|---|---|---|---|
| [32] | In this paper, a deep model is proposed for traffic flow forecasting. This model consists several convolution layer and nested LSTM [26]. The input of this model is a image of the network link. This image shows the average speed in each link | Dropout | No |
| [33] | Transportation Mode Detection and use an ensemble of CNNs for classifying transportation mode to five classes. The input of the deep model is a tensor of GPS preprocessed data. | Dropout, Early Stopping, and Data Augmentation | Yes |
| [34] | Using a CNN to detect the crack of roads. The input is of the deep model is the images of roads. | Two approaches are used. The first one is dropout. The second one is that feed the one input with different views and compute a probability between the outputs. | No |
| [35] | The paper considered the different ITS applications including image processing, transportation mode detection and driving evaluation datasets. | Using adaptive regularization in deep networks including adaptive dropout and adaptive weight decay. | Yes |

## 4- Driver Style Evaluation based on Sensors of Smartphone

In this part, we propose a driver style evaluation model based on the deep network. Also, we compare the performance of the model with regularization and without it.

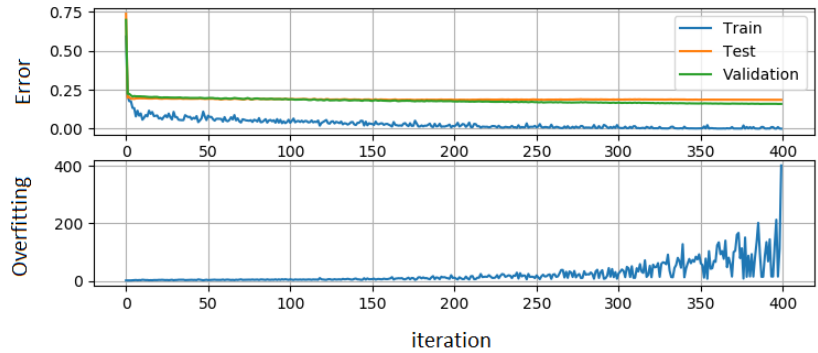

*Figure 1: The training procedure error and the value of overfitting without any regularization on Driver Style Evaluation*

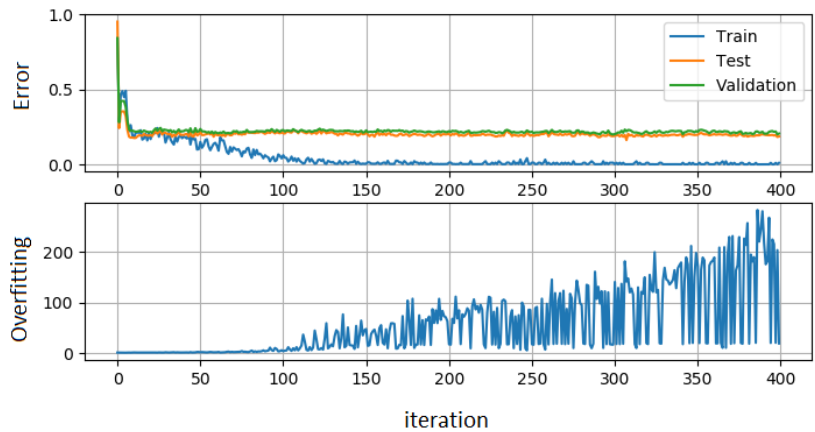

*Figure 2: The training procedure error and the value of overfitting with dropout as regularization*

**4-2 The Results of Driver Style Evaluation**

After augmentation, we consider two CNNs as the model to classify the dangerous driver and normal driver. In first model, we do not use any regularization schemes, but in the second model, we use dropout. As one can see in figures (1) and (2), we illustrate the performance of the each model in training procedure. The figures are shown two criterial error and overfitting. The overfitting is calculated based on following equation:

$$v(t) = \frac{Error_{Validation}}{Error_{Train}}, \tag{5}$$

where $Error_{Validation}$ and $Error_{Train}$ indicate to error of the model on the validation and the train dataset, respectively. The final result for CNN without regularization is 88% accuracy and for CNN with regularization is 93% accuracy.

**4- Conclusion**

The deep models have dramatically high performance for complex problems, but in many real cases, they have a big problem named overfitting. This problem causes that the model has not suitable performance on the test data. There are many solutions to solve this problem, which are discussed in this paper. In addition, in data science, we use different types of models to predict unknown inputs. ITS uses data science to investigate the environments such as transportation mode detection, driving style evaluation, or traffic flow prediction. These problems are very complex. Then, experts cannot extract all of the important features from the raw data. Therefore, in this case, we have to use the deep models. However, in some applications, the collecting high amount of the data is not possible or the data is collected imbalance. These cause that the model becomes over-fitted. We survey some of the important papers, which deal with the overfitting problem in ITS application. In addition, to show the importance of solving the overfitting problem and using deep models in ITS, we test CNN models on driver style evaluation.

**6-References**